\pdfoutput=1

\documentclass[11pt]{article}

\usepackage[]{ACL2023}

\usepackage{times}
\usepackage{latexsym}

\usepackage[T1]{fontenc}

\usepackage[utf8]{inputenc}

\usepackage{microtype}
\usepackage{adjustbox}
\usepackage{graphicx}
\usepackage{booktabs}
\usepackage{multirow}
\usepackage{adjustbox}
\usepackage{listings}
\usepackage{caption}
\usepackage{xspace}
\usepackage{amsfonts}
\usepackage{amsmath}
\usepackage{colortbl}
\usepackage{array}
\usepackage{ifthen}
\usepackage{tabularx}

\RequirePackage{type1cm}
\RequirePackage{color}
\RequirePackage{soul}
\setstcolor{blue}
\definecolor{violet}{rgb}{0.5,0.0,0.5}
\newsavebox\bscombox
\newcommand{\bscom}[3][]{%
	\sbox{\bscombox}{\fontsize{8}{9}\selectfont#1#2#3}
	\noindent
	\st{#2}{\selectfont
		\color{blue}#3\ifx\\#1\\\else{\fontsize{8}{9}\selectfont\color{violet}[#1]}\fi
	}
}

\usepackage[table,xcdraw]{xcolor} 
\usepackage{inconsolata}

\title{LLMs as Science Journalists: Supporting Early-stage Researchers in Communicating Their Science to the Public}

\author{
	\textbf{Milad Alshomary \textsuperscript{\dag}},
	\textbf{Grace Li\textsuperscript{\S}},
	\textbf{Anubhav Jangra\textsuperscript{\dag}},
	\\
    \textbf{Yufang Hou\textsuperscript{\ddag}},
	\textbf{Kathleen McKeown\textsuperscript{\dag}},
     \textbf{Smaranda Muresan\textsuperscript{\dag}}
	\\
	\\
	\textsuperscript{\dag}Columbia University, New York, USA
	\\
	\textsuperscript{\ddag}IT:U Interdisciplinary Transformation University Austria
	\\
    \textsuperscript{\S}University of Chicago, Illinois, USA
    \\
	\small{
		\textbf{Correspondence:} \href{ma4608@columbia.edu}{ma4608@columbia.edu}
	}
}

\begin{document}
\maketitle

\begin{abstract}
The scientific community needs tools that help early-stage researchers effectively communicate their findings and innovations to the public. Although existing general-purpose Large Language Models (LLMs) can assist in this endeavor, they are not optimally aligned for it. To address this, we propose a framework for training LLMs to emulate the role of a science journalist that can be used by early-stage researchers to learn how to properly communicate their papers to the general public. We evaluate the usefulness of our trained LLM Journalists in leading conversations with both simulated and human researchers.  
Our experiments indicate that LLMs trained using our framework ask more relevant questions that address the societal impact of research, prompting researchers to clarify and elaborate on their findings. In the user study, the majority of participants who interacted with our trained LLM Journalist appreciated it more than interacting with general-purpose LLMs.
\end{abstract}

\section{Introduction}
\label{sec:intro}

As the communication gap between the scientific community and the public grows \cite{hunter2016communications}, a substantial body of research is examining ways to address this and restore public confidence in science. One direction is to support early-stage researchers in communicating their research effectively to the public. This is a challenging task due to the overly technical content of their work and a lack of a comprehensive understanding of the field's broader societal context. While a staggering amount of resources are available to learn how to frame one's research \cite{padt2023teaching}, they are often presented as how-to guidelines and are ineffective for learning. The rise of AI assistants can offer an interactive alternative for learning. For example, researchers can use LLMs to simulate a lay audience with whom they can converse to practice communicating their paper. Nevertheless, general-purpose AI assistants such as Llama-3 or GPT-4 are not specifically aligned for this goal. It is the researcher's responsibility to figure out how to best use the AI assistant to build communication skills, and users, when presented with an AI assistant, often rely fully on it to complete the task, which diminishes the learning effect \cite{kosmyna2025your}.

\begin{figure}
    \centering
    \includegraphics[]{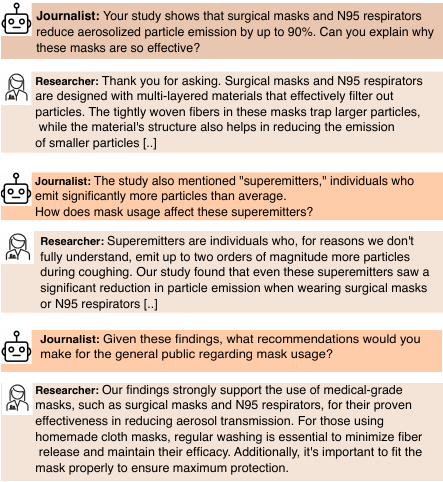}
    \caption{Example conversation on a paper by \citet{asadi2020efficacy}, as simulated by {\texttt Deepseek-R1} model guided by this \href{"https://medicalxpress.com/news/2020-09-surgical-n95-masks-block-particles.html"}{press release.}}
    \label{fig:example_conv}
\end{figure}

We hypothesize that an AI-guided conversation, where AI takes the role of a science journalist, is more effective for researchers to learn how to communicate their scientific findings compared to general-purpose AI systems, where the researcher must lead the conversation, attempting to determine the best learning strategy. As demonstrated in Figure \ref{fig:example_conv}, in an AI-guided conversation, the Journalist LLM will ask the researcher questions about their paper, prompting them to consider the aspects of their research that are relevant to the public. To this end, we focus on three important aspects of good science communication: (1) communication should be in an accessible language to the public \cite{lang2025jargon}, (2) it should emphasize the societal impact of the research and its findings \cite{kreissl2015societal}, and (3) it should put the research in its scientific context \cite{lowe2002complexities}. 

To test our hypothesis, we design a novel framework to automatically synthesize data for training LLMs to take the role of science journalists. In particular, due to the lack of science conversation datasets and the assumption that press releases serve as a reliable mechanism for communicating science, we start from a corpus of scientific papers and their corresponding press releases \cite{pu2024scinews}, and use an oracle LLM to simulate scientific conversations on all papers, aligned with their corresponding press release. We then use these synthesized conversations to train an LLM through a two-phase training procedure (supervised fine-tuning followed by preference learning), enabling it to play the role of a journalist\footnote{Throughout the paper, journalist and science journalist are used interchangeably.}. 

To evaluate the quality of our trained LLM journalists, we designed two evaluations. In the first, we simulate conversations between the evaluated LLM journalist and an LLM researcher (prompted to act as such) on unseen papers and evaluate the quality of the LLM journalist's questions. We train two LLMs (\texttt{Llama-3-8B} and \texttt{Qwen2.5-7B}) using our training framework and compare their qualities against their corresponding vanilla equivalents, ablated versions, and closed-source baselines. Our results demonstrate that our fine-tuned LLMs facilitate better-quality conversations, particularly in discussing the societal impact of the research, its scientific context, and using accessible language to a broader audience. Second, we design a user study in which computer science PhD students are tasked with writing a lay summary of their research with the help of a general-purpose AI or our fine-tuned LLM journalist. In particular, each participant (1) converses with the evaluated system for 10 minutes, (2) writes a lay summary, and (3) answers a set of questions about their experience. This procedure is repeated for each system on two different research papers authored by the participant. Finally, the subject answers a set of questions to compare the two interaction experiences. Our results show that the majority of participants favored the fine-tuned LLM journalist, citing its usefulness in prompting them to recall relevant content about their paper and to better frame it within its societal and scientific context.
In summary, our contributions are: 
\begin{itemize}
    \item A training framework to align LLMs to be better at assisting researchers in communicating their findings to the public.
    \item A synthesized corpus of high-quality scientific conversations. 
    \item Evidence that our fine-tuned LLM journalists are better than general-purpose counterparts in guiding researchers to better communicate their findings to the public.
\end{itemize}
We make our code, the synthesized dataset, trained models, and the chat interface publicly available.\footnote{To test our LLM Journalist, upload your paper into our \href{https://radiotelephonic-calyciform-lorilee.ngrok-free.dev/}{interface} and select system "Mango" from the drop-down list.}

\begin{figure*}
    \centering
    \includegraphics[]{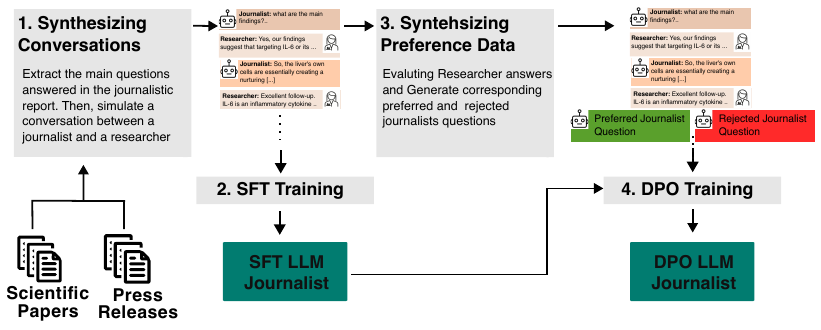}
    \caption{Our training framework: Starting with a corpus of scientific papers and their press releases, we use \texttt{Deepseek-R1} to synthesize conversations. Second, we use supervised fine-tuning (SFT) to train an LLM to act like a journalist. Third, we synthesize preference data favoring follow-up and societal questions. Finally, the preference data is used to perform preference learning (DPO) over the SFT LLM.}
    \label{fig:training-framework}
\end{figure*}

\section{Related Work}
Science journalism is about communicating complex scientific findings to a broader, non-technical audience with the goal of bridging the gap between science and the general public, thereby increasing public trust in science \cite{hunter2016communications}. Among many topics, educating researchers on how to frame their findings with respect to the general public is a well-studied theme \cite{swords2023science, reincke2024identifying}. For example, \citet{fick2025talk} proposed and evaluated a training program to improve graduate students’ science communication skills. While we share similar goals, our work aims to provide researchers with tools that enable them to learn the skills in an interactive and practical manner.

In the field of natural language processing (NLP), several related works have investigated approaches for automatically summarizing scientific papers into journalistic reports \cite{cardenas2023don, cachola2020tldr, pu2024scinews}. These efforts resulted in the collection of corpora of scientific papers and their corresponding journalistic reports. \newcite{cardenas2023don} proposed a discourse-guided approach to summarizing scientific research. Our goal is to provide researchers with tools that help them practice communicating their work to the public, rather than fully automating the process. Nevertheless, we use their produced corpora to synthesize conversations that we train our LLMs on.

Large language models (LLMs) have been increasingly used as tutors in education \cite{xu2024large}. For example, \citet{vanzo2025gpt} shows that prompting GPT-4 to act like a tutor produced measurable learning benefits and engagement gains. \citet{scarlatos2025training} propose a framework for training LLMs to act like tutors. Similarly, we are interested in examining the benefit of training LLMs to assist researchers in communicating science to the public by playing the role of a journalist rather than a tutor.
\section{AI for Scientific Conversations}
\label{sec:approach}
We aim to develop and test an AI-guided conversational system that serves as a journalist, prompting researchers to communicate their research findings more effectively. We will first present the key aspects of what we consider effective communication, and then propose our framework for aligning LLMs to serve as assistants for researchers. 

\subsection{Aspects of Good Science Communication} \label{good-sci-comm}
We consider three main aspects one should emphasize when communicating scientific research to the public: The societal impact, the scientific context, and the accessibility of the language used.

\paragraph{Societal Impact} To improve public trust and confidence in science, one has to communicate how relevant the given research is to social issues that the public cares about \cite{cologna2025trust}. We follow \citet{kreissl2015societal} by defining the social impact as "\emph{the changes that occur in people, communities, and/or environments outside of academia as a result of the research process or findings. These impacts can include conceptual changes, such as new perspectives on socio-ecological challenges, as well as instrumental and capacity-building changes that lead to longer-term social and environmental impacts over time. Stakeholder engagement plays a significant role in driving these societal impacts}".

\paragraph{Scientific Context} Situating research findings within their broader scientific context enables the public to better appreciate the robustness and cumulative nature of the underlying evidence, as well as to recognize related research on the same topic, which contributes to the mitigation of misinformation \cite{lowe2002complexities}. Moreover, providing this contextualization facilitates a clearer understanding of the research’s degree of novelty and its specific contribution to the existing body of knowledge.

\paragraph{Language Accessibility} Communicating research findings should be achieved by using straightforward and simple language that is free from technical jargon. Accessibility considers not only the use of simple vocabulary \cite{kincaid1975derivation} but also the use of rhetorical devices, such as examples and analogies, that enhance readers' understanding. Accessibility has been extensively studied in related work and is shown to be crucial for public communication \cite{lang2025jargon}.

\subsection{Training Framework}
We hypothesize that a specialized LLM designed to play the role of a science journalist interacting with researchers is more effective than a general-purpose one in supporting researchers in building science communication skills. To validate this hypothesis, we need to equip LLMs with such a skill. To this end, and due to the lack of science conversation datasets, we synthesized our own corpus of conversations, starting from press releases that serve as a reliable proxy for good communication of papers, albeit in monological form. We then use this corpus to train LLMs to act as journalists. In the following, we detail the components of our framework as sketched in Figure \ref{fig:training-framework}, including the LLM as a Judge setup, data synthesis, and the training process. 

\paragraph{LLM as a Judge}
We aim to assess various quality aspects of human-written press releases, synthesized conversations, and generated questions by the LLM journalist. To achieve this, we employ an LLM as an evaluator and create specific prompts tailored to each quality aspect. Specifically, we utilize {\texttt{DeepSeek-R1-Distill-Qwen-32B}}\footnote{\url{https://huggingface.co/deepseek-ai/DeepSeek-R1-Distill-Qwen-32B}} as the LLM evaluator, with our evaluation prompts detailed in the appendix. The LLM judge is then used in steps 1 and 3 of our framework to ensure the quality of press releases and to score the quality of the researcher's answers.

\paragraph{1. Synthesis of Scientific Conversations.} To our knowledge, there exists no direct access to scientific conversations over research papers. Therefore, we follow a quality-guided approach to synthesize such a conversational corpus. Specifically, we begin with a corpus of scientific papers and their corresponding press releases, which were made public by \citet{pu2024scinews}. The corpus consists of around 41k papers and press releases covering up to nine scientific domains (Medicine, Biology, Physics, Computer Science, etc.). The corpus is split into training (80\%), validation (10\%), and test set (10\%). To ensure high-quality press releases, we run the training split's press releases through our LLM judge mentioned above and generate quality scores for each press release, rating how much it discusses the societal impact and scientific context, as well as its language accessibility. Prompts \ref{lst:societal_impact_ques_extraction}, \ref{lst:scientific_context_ques_extraction}, and \ref{lst:clarity_ques_extraction} present the reasoning behind the scoring as explained to the LLM Judge. We then retain only instances with press releases that have an accessible language score higher than three and an average score of scientific context and societal impact higher than two. The resulting training corpus consists of 18k pairs of scientific papers and their press releases. Next, considering these press releases as ground truth for effective communication of scientific papers, we prompt {\texttt{DeepSeek-R1-Distill-Qwen-32B}} to synthesize a scientific conversation between a researcher and a journalist for each research paper, aligned with the corresponding press release. Prompt \ref{lst:conv_synth_prompt} shows the exact prompt used to generate the conversations. The input includes the first 1k tokens of the scientific paper and the full press release (averaging 813 tokens). The resulting conversation dataset averages 12 turns, with journalist questions averaging 28 tokens and researcher answers averaging 70 tokens. Figure \ref{fig:example_conv} shows an example conversation.


\paragraph{2. Supervised Fine-tuning (SFT).} Given a dataset of scientific papers and the corresponding synthesized scientific conversations, we perform supervised fine-tuning of an LLM to produce the journalist utterances. From the 18k scientific conversation corpus, we distill around 80k journalist utterances, each with its own dialogue history and input paper. We split this corpus into a training and validation split and train two large language models for 3 epochs: \texttt{Llama-3-8B} and \texttt{Qwen2.5-7B} to ensure that our hypothesis holds for different LLMs. More details on the training can be found in Section \ref{app:training-framework} in the appendix. We refer to the two resulting models as \texttt{SFT-Llama} and \texttt{SFT-Qwen}.

\begin{table*}[t!]
\setlength\tabcolsep{4pt}
\centering
\begin{tabular}{lcccccc}
& \multicolumn{4}{c}{\bf LLM as a Judge} & \multicolumn{2}{c}{\bf Automatic Eval}\\
\cmidrule(l){2-5} \cmidrule(l){6-7} 
& \textbf{Access.} $\uparrow$ & \textbf{Scientific.} $\uparrow$& \textbf{Societal.} $\uparrow$ & \bf AVG. $\uparrow$& \textbf{Redund.} $\downarrow$& \textbf{Follow.} $\uparrow$\\
\midrule
\bf GPT-4o-mini (Simple) &  \underline{0.65} & \underline{0.43} & 0.21 & \underline{0.35} & 0.75 & 0.58 \\
\bf GPT-4o-mini (Advanced) &  0.58 & 0.13 & \bf \underline{0.41} & 0.25 & \underline{0.74} & \bf \underline{0.65} \\
\midrule
\bf Qwen (Simple) &  0.64 & 0.41 & 0.07 & 0.16 & 0.71 & 0.61 \\
\bf Qwen (Advanced)& 0.69 & 0.45 & 0.16 & 0.30 & 0.77 & \textbf{\underline{0.65}}\\
\bf SFT-Qwen &  0.57 & 0.34 & 0.19 & 0.30 & \underline{\textbf{0.62}} & 0.51 \\
\bf DPO-Qwen & \bf \underline{0.85} & \bf \underline{0.60} & \underline{0.23} & \bf  \underline{0.42} & 0.70 & 0.55\\
\midrule
\bf Llama (Simple) & 0.63 & 0.32 & 0.15 & 0.26 & 0.76 & 0.33 \\
\bf Llama (Advanced) &  0.68 & 0.31 & 0.12 & 0.23 & 0.78 & 0.56\\
\bf SFT-Llama & 0.67 &  \underline{0.34} & 0.18 & 0.30 & 0.66 & 0.48\\
\bf DPO-Llama & \underline{0.81} &  0.33 & \underline{0.23} & \underline{0.35} & 0.77 & 0.56\\
\bottomrule
\end{tabular}
\caption{The evaluation results of the LLM journalists (SFT-* and DPO-*) compared to the prompted correspondents and gpt-4o mini, in terms of percentage of questions about the accessibility (Access.), Scientific context (Scientific.), Societal impact (Societal), and their harmonic average (AVG.), as well as automatic measures of redundancy (Redund.) and Follow-up (Follow.). Highlighted in bold is the best performance overall, and underlined is the best value among same family of models.}
\label{table:automatic-eval-results}
\end{table*}

\paragraph{3. Preference Learning.} While the previous SFT training step helps LLMs mimic research journalists, upon initial inspection, we noticed that the SFT-trained models don't often ask follow-up questions to clarify the researcher's answers or questions about the societal impact. Therefore, as a next step, we create a preference dataset and further train SFT models to enhance journalist-like behavior, focusing on: (1) follow-up questions for clarity, and (2) questions about societal research impact. In particular, we sample 19k researcher answers from the synthesized conversation dataset, along with the conversation history and input paper, and create the preference data as follows. For each of the instances in the dataset, we prompt the LLM judge to evaluate two factors: (1) if the utterance is unclear, and (2) if it includes any highly technical concepts (Prompts \ref{lst:follow-up-vague-answer} and \ref{lst:follow-up-complex-concepts} in the appendix). If the researcher's utterance has any of these issues, the SFT model is then prompted to create a follow-up clarification question, which is chosen as the accepted journalist question, and a general question, which is designated as rejected. Otherwise, the SFT model is prompted to generate a question about the societal impact of the paper, which is considered the preferred question. All prompts used for synthesizing this dataset are in Figure \ref{fig:dpo-generation-prompts} in the appendix.
We then train each of the SFT models, \texttt{SFT-Llama} and \texttt{SFT-Qwen}, for one epoch using the DPO training algorithm \cite{rafailov2023direct}. The resulting two models are then referred to as \texttt{DPO-Llama} and \texttt{DPO-Qwen}. More details on the training can be found in Appendix \ref{app:training-framework}.
\section{Automatic Evaluation}
\label{automatic-eval}
In this section, we first present our experiments to automatically evaluate our fine-tuned LLM-Journalist against their corresponding general-purpose LLMs, prompted to act as journalists.

\subsection{Experiment Setup}

\paragraph{Simulating Conversations} To evaluate the performance of the evaluated LLM Journalists, we sample 500 instances from the test set. For each instance, we simulate a conversation between the evaluated LLM, playing the role of the journalist, and the vanilla \texttt{Llama-3-8B} model, playing the role of the researcher. In particular, we first pass the paper (title and the first 1k tokens covering the abstract and the introduction) to the journalist LLM to generate the initial question, and then we pass the paper along with this question to the LLM researcher to obtain an answer. We repeat this process automatically for five rounds to build a conversation (10 utterances). 
\paragraph{Baselines} To evaluate the value of our training framework, we compare our trained models \texttt{DPO-Llama} and \texttt{DPO-Qwen} against their ablated versions (SFT-Llama and SFT-Qwen), and the corresponding vanilla versions, which are prompted using a simple (Prompt \ref{lst:journalist_role_simple}) and a more elaborate (Prompt \ref{lst:journalist_role_advanced}) prompts that inform the LLM how to act as a journalist. Additionally, we compare against \texttt{GPT-4o-mini} as an example of a closed-source baseline model also prompted using the simple and advanced prompts.

\paragraph{Evaluation Measures} For evaluation, we quantify the quality of the questions asked by the LLM journalist as follows. We design three prompts (Figure \ref{fig:eval-quality-prompts} in the appendix) to extract questions that address societal impact, scientific context, and seek clarification, given a conversation. The corresponding evaluation measure, then, is the percentage of these questions from the total number. Ideally, a good LLM journalist would ask a well-balanced distribution of questions covering the three aspects. Furthermore, we evaluate the redundancy in the LLM journalist's questions across conversations and the percentage of follow-up questions they ask. Redundancy is estimated using a self-referenced redundancy metric \cite{chen2021training} across all journalist questions within a conversation, and follow-up behavior is measured as the average similarity between each generated question and the researcher’s preceding response. Both metrics rely on sentence-level representations derived from the large Sentence-BERT NLI model \cite{reimers-2019-sentence-bert}.

\subsection{Results}
Table \ref{table:automatic-eval-results} presents the evaluation results. Overall, in terms of the harmonic mean of the three types of questions (AVG. column in the table), both DPO-Llama and DPO-Qwen achieve the highest scores of 0.35 and 0.41, respectively, striking a strong balance. While the SFT-trained models lead to a better percentage of societal questions (Societal.), they don't ask many clarifying questions (Access.) compared to the baselines. On the contrary, as intended, the DPO training increased both the percentage of clarifying and societal impact questions compared to all baselines, resulting in the best harmonic average score. Moreover, we observe that the advanced prompt led to an improvement in the performance of the Qwen model under all three aspects (0.30 compared to 0.16), but negatively impacted the performance of the Llama and GPT-4o-mini models (0.23 compared to 0.26, and 0.25 compared to 0.35). Overall, the results suggest that prompting LLMs, even closed-source ones, is insufficient to achieve strong performance, underscoring the importance of our training framework in aligning LLMs to emulate the style of science journalists.

In terms of Redundancy and Follow-up, we observe a clear trade-off between the two measures. The advanced prompt reliably enhances follow-up quality, yielding more responsive and context-aware questions. However, this also leads to higher redundancy, suggesting that its stronger scaffolding encourages repetitive patterns. In contrast, the SFT-trained journalists achieve the lowest redundancy within each family, indicating more diverse questioning, yet they produce weaker follow-up questions, implying a loss of conversational grounding. DPO-trained models strike a middle ground, as they improve follow-up behavior relative to SFT while maintaining redundancy levels comparable to the prompted baselines. 

\section{User Study} \label{user-study}
To assess the usefulness of our system, we design a user study to compare two different interaction paradigms to assist early-stage researchers in communicating their research to the public: (1) the general-purpose AI, where the researcher leads the conversation, and (2) the AI-guided conversation, where our fine-tuned LLM plays the role of a journalist asking questions about the paper.

\subsection{User Study Setup}

\paragraph{Evaluated LLMs} Given the automatic evaluation results (Table \ref{table:automatic-eval-results}), we observe that DPO-Qwen performs the best, but in terms of vanilla performance, the Llama model (Simple Prompt) outperforms Qwen (Simple Prompt). Therefore, to have a stronger baseline, we use vanilla Llama-3 as the LLM for general-purpose interaction and its corresponding \texttt{DPO-Llama} for the AI-guided paradigm. The vanilla Llama-3 model serves as a useful assistant that answers any questions the researcher poses. Similar to the automatic evaluation, both LLMs receive the introduction of the paper (up to 1000 tokens) as an input to converse about.

\paragraph{Chat Interface}
To facilitate interaction with the LLMs, we design a chat interface using the Streamlit framework and serve the LLMs through the vLLM library \cite{kwon2023efficient}. We also make this interface publicly available to use\footnote{Link to the \href{https://radiotelephonic-calyciform-lorilee.ngrok-free.dev/}{interface}}.  

\paragraph{User Study Design}
Given our focus on supporting early-stage researchers in communicating their science to the public, we recruit current PhD students because they are actively developing both their research identities and their communication practices. We design a between-subjects study to compare each participant's experiences of both systems (LLM Journalist and general-purpose LLM). Participants experience the systems in a counterbalanced order to reduce order effects on the results. 

Participants bring two of their own research papers, which they upload to the system. Before the study, one of the authors provides an introduction to the principles of effective science communication to ensure participants understand the task and what effective public-facing communication entails (see Section \ref{good-sci-comm}). Each participant initially interacts with the designated LLM chat for 10 minutes. Then, they spend 10 minutes composing a lay summary of their paper. Participants can copy/paste excerpts of their own paper, but not their conversation history with the LLM. Afterwards, they fill out a system survey that covers questions on the usefulness of the given system and interaction in helping users remember the most relevant details of their paper (Relevance), use everyday language (Access), frame the paper with respect to related work (Scientific), motivate the societal impact (Societal), and whether the follow-up questions were relevant to the context (Follow-up). After interacting with both systems, participants complete a comparative survey to compare their experiences with both systems, answering questions that cover which system was most helpful in helping them recall details about their paper, was easier to communicate with, and overall, the system they prefer using in their daily life (see Appendix \ref{sec:user_study_questions} for all survey questions). 
Due to the time-demanding nature of the task (1 to 1.5 hours per participant), we succeeded in recruiting only 9 PhD students in the field of computer science as participants, each was compensated with \$45. Details on the recruitment process and example conversations in Appendix \ref{sec:user_study_questions}.

\begin{figure}[t!]
    \centering
    \includegraphics[scale=0.9]{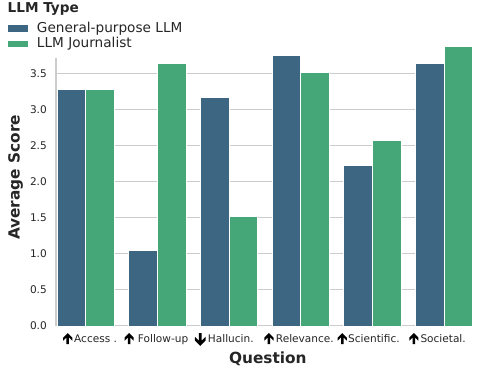}
    \caption{Aggregation of participants' scoring of their experience interacting with each of the Systems}
    \label{fig:survey_1_questions}
\end{figure}

\subsection{Results}

\subsubsection{Analysis of User Feedback}

We aggregate participant scores for each system survey to understand user ratings of the various evaluated aspects. Figure \ref{fig:survey_1_questions} shows the average scores achieved by each of the two evaluated systems as reported by the participants for each of the evaluation questions. First, as for the system's usefulness in helping participants frame their papers in terms of {societal impact (Societal)}, {scientific context (Scientific)}, and {using everyday language (Access)}, our LLM Journalist achieves equal or higher scores than the general-purpose equivalent (3.7 vs. 3.4 for societal impact, 2.4 vs. 2.1 for scientific context, and 3.1 on accessibility for both systems). Second, our system produced fewer hallucinations about the paper compared to the baseline, which is expected, as our system only asks relevant questions about the paper and converses with the participant, whereas the baseline is often prompted by the participants to produce summaries on the paper. Finally, users found the follow-up questions (Follow-up) of our system to be more useful.
As for the comparative survey, our results show:
\begin{itemize}
    \item 77\% of the participants preferred the LLM journalist to learn communication skills.
    \item As for ease of communication and the preferred lay summary, 66\% of participants selected our LLM journalist.
    \item 66\% participants agreed or strongly agreed that their interaction with our LLM journalist was more helpful for recalling details about their paper when writing the summary.
\end{itemize}

\paragraph{Qualitative Analysis of User Feedback}

To better understand the affordances and challenges of the LLM journalist system, we conduct inductive thematic analysis on the open-ended survey questions \cite{clarke2014thematic} (see Appendix \ref{codebook} for the codes used). In terms of \textbf{affordances of LLM journalist system}, 7 participants identified several affordances of the LLM journalist system. First, the LLM journalist \textit{reduced prompting burden} by lowering the cognitive effort required to initiate and manage interactions (n=3). Participants reported that being prompted with questions made the interaction feel less mentally taxing than general-purpose prompting. For P8, using the LLM journalist system was ``\emph{less cognitively tiring (it prompted me with questions so I did not have to think about how to prompt the system).}" In contrast, when P7 was using the baseline system, they remarked that ``\emph{I had to keep reminding the model of the constraints (i.e., writing for a lay audience) that made the interaction slower and more frustrating.}'' 
Second, the LLM journalist \textit{encouraged active engagement} by requiring participants to respond to questions rather than passively accepting an LLM-generated writing (n=3). Participants described thinking more critically about their arguments, generating new ways to talk about their papers, and internalizing key points through repeated articulation, in contrast to vanilla interactions that felt more surface-level and less cognitively engaging. For P6, ``\emph{[LLM journalist] made me think deeper about the points I was making... [the baseline] was more passive and wasn’t really making me think too much.}" Finally, the LLM journalist supported \textit{perspective-taking for a lay audience} by reminding participants of the audience, asking them to explain technical jargon or concepts (n=2). Participants noted that the system more closely resembled real-world communication contexts, such as pitching their work to non-experts, and made it easier to frame their research for a general audience. While the LLM journalist system was helpful, participants also reported challenges, including misaligned questions with their interests (n=1), narrow question coverage (n=1), and redundant questions (n=2).

\begin{figure}
    \centering
    \includegraphics[width=\linewidth]{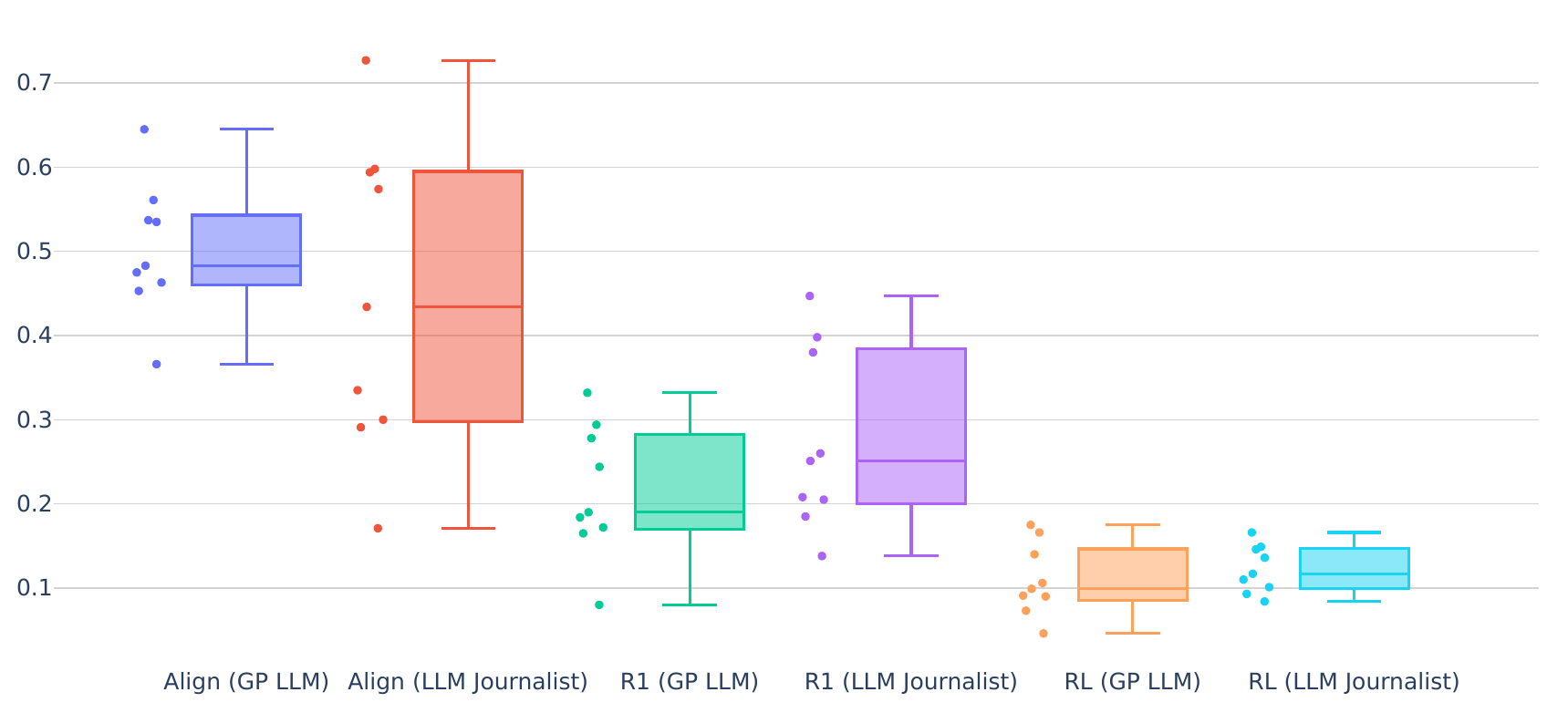}
    \caption{Statistics of information overlap between the user-written lay summary and their system interactions, computed using AlignScore (Align.) \cite{zha2023alignscore}, ROUGE-1 (R1) and ROUGE-L (RL) \cite{lin2004rouge}. }
    \label{fig:manual_summary_eval}
\end{figure}

\subsubsection{Lay Summaries and Interactions} 
Although the study is limited in scale (N = 9), we can still draw several observations from the artifacts obtained from the user study, namely the participant-written lay summaries, their interactions with the system, and the alignment between the two. We observe from Figure \ref{fig:manual_autoeval} that the LLM journalist displays a higher follow-up score and a tighter inter-quartile range of redundancy with comparable scores to the baseline.  
In terms of overlap, lay summaries written with the help of the LLM journalist have higher unigram overlap compared to general-purpose LLM, as indicated by a 28\% difference in ROUGE-1 score (Figure \ref{fig:manual_summary_eval}). This higher word retention from the system interaction, in lay summary, along with a more concise interaction as depicted by shorter conversation length, indicates a higher engagement quality with the LLM journalist compared to the general-purpose LLM. Moreover, Figure \ref{fig:manual_numwords} illustrates the semantic influence of the length of user interaction (measured using \#words) on the lay summaries, computed using AlignScore \cite{zha2023alignscore} between the lay summaries and system interactions. The plot suggests a systematic difference in how the two system interactions influence the paper summaries. The general-purpose LLM interactions are generally longer than those of the LLM journalist, and notably, this interaction length does not uniformly translate into higher alignment. In contrast, the LLM journalist exhibits diverging content overlap between the summaries and the conversations, with similarly sized conversations yielding an AlignScore ranging from low alignment of 0.17 to very high alignment of 0.73.

\paragraph{Qualitative Analysis of User Interactions} 

To understand the patterns in the user interaction with the two systems, we conduct an inductive thematic analysis of each user-issued message for both systems (LLM journalist and baseline). Across systems, there are 3 broad categories of prompt types: (1) \textit{directive}, when the user issues an instruction or action for the system to perform, (2) \textit{answering a question}, when the user responds to a question that the system posits, and (3) \textit{asking question} when the user seeks to gain information from the system (Table \ref{tab:user-interaction-codes} in appendix). For the Baseline system, we found that most interactions had a majority--at least 50\% of user-issued messages--were directive prompts (6 of 9 interactions), offloading the task to the assistant. On the contrary, for the LLM journalist system, all participants (9 of 9) had the majority of their interactions as answering questions posed by the system. This places the participant in the position to practice translating and summarizing their research. More insights can be found in Section \ref{sec:analysis_of_user_interactions}.
\section{Conclusion}
We emphasize the significance of AI in enabling early-stage researchers to effectively communicate their work to the public. To this end, we propose a framework that trains LLMs to act as journalists, prompting researchers to situate their papers in societal and scientific contexts and asking questions to clarify missing or complex details. To overcome the scarcity of scientist–journalist dialogues, our framework synthesizes such conversations from existing journalistic reports and uses them for training. We test whether this training produces better models than simple prompting and whether a journalist-style LLM is more effective than general-purpose AI assistance. Automatic evaluation shows that our trained LLM journalist outperforms a prompt-based correspondent by asking a more balanced set of questions across multiple quality dimensions. In the user study, most participants preferred interacting with our LLM journalist over using a general-purpose AI assistant.
\section{Limitation}
Despite the present evidence demonstrating the value of our training framework in aligning LLMs to be better science journalists, we would like to still highlight some limitations. First, we focused our evaluation criteria on covering journalists' questions about the scientific context, societal impact, and whether they prompted the researcher to clarify their answers. However, our evaluation can be extended to other dialogue quality criteria to have a more comprehensive evaluation framework. Second, so far, we have assumed that we have one single themed public audience. A more effective framework for learning science communication can consider varying audiences with different backgrounds, knowledge, and goals. The user can then define such an audience to customize the LLM journalist with such knowledge, leading to questions that are specifically interesting for that given audience. Third, we evaluated generated conversations of 10 turns only; a thorough analysis of longer conversations can give better insights into how the evaluated models behave as the length of the conversations increases. Fourth, our best-performing trained models (DPO-Qwen and DPO-Llama), although they excel at asking questions about societal impact and clarification, fall behind in addressing questions about positioning the research in the correct scientific context. However, using our framework, one can modify the preference data prompts and design other quality criteria that align with the trained LLM journalists accordingly. Finally, our user study is limited in size. Despite the clear preference of our participants for the designed LLM journalist, a bigger participant sample can allow a more thorough analysis. For example, diversifying the backgrounds of participants, their career levels, or their disciplines can lead to more informative insights.

Regarding ethical concerns, we would like to highlight that our trained models serve as journalists, asking questions about the paper, which reduces the potential for hallucinations or harmful content. Their role is limited to prompting the user with appropriate cues and questions to guide them in communicating their research.


\bibliography{anthology,custom}

@inproceedings{pu2024scinews,
  title={SciNews: From Scholarly Complexities to Public Narratives--a Dataset for Scientific News Report Generation},
  author={Pu, Dongqi and Wang, Yifan and Loy, Jia E and Demberg, Vera},
  booktitle={Proceedings of the 2024 Joint International Conference on Computational Linguistics, Language Resources and Evaluation (LREC-COLING 2024)},
  pages={14429--14444},
  year={2024}
}

@article{rafailov2023direct,
  title={Direct preference optimization: Your language model is secretly a reward model},
  author={Rafailov, Rafael and Sharma, Archit and Mitchell, Eric and Manning, Christopher D and Ermon, Stefano and Finn, Chelsea},
  journal={Advances in neural information processing systems},
  volume={36},
  pages={53728--53741},
  year={2023}
}

@article{cologna2025trust,
  title={Trust in scientists and their role in society across 68 countries},
  author={Cologna, Viktoria and Mede, Niels G and Berger, Sebastian and Besley, John and Brick, Cameron and Joubert, Marina and Maibach, Edward W and Mihelj, Sabina and Oreskes, Naomi and Sch{\"a}fer, Mike S and others},
  journal={Nature Human Behaviour},
  pages={1--18},
  year={2025},
  publisher={Nature Publishing Group UK London}
}

@article{kreissl2015societal,
  title={Societal impact assessment},
  author={Kreissl, Reinhard and Fritz, Florian and Ostermeier, Lars},
  year={2015},
  publisher={Elsevier}
}

@article{lowe2002complexities,
  title={The complexities of communicating science},
  author={Lowe, Ian},
  journal={TheAustralian Universities' Review},
  volume={45},
  number={2},
  pages={3--22},
  year={2002},
  publisher={National Tertiary Education Union South Melbourne}
}

@article{lang2025jargon,
  title={Jargon and Readability in Plain Language Summaries of Health Research: Cross-Sectional Observational Study},
  author={Lang, Iain A and King, Angela and Boddy, Kate and Stein, Ken and Asare, Lauren and Day, Jo and Liabo, Kristin},
  journal={Journal of Medical Internet Research},
  volume={27},
  pages={e50862},
  year={2025},
  publisher={JMIR Publications Toronto, Canada}
}

@inproceedings{chen2021training,
    title = "A Training-free and Reference-free Summarization Evaluation Metric via Centrality-weighted Relevance and Self-referenced Redundancy",
    author = "Chen, Wang  and
      Li, Piji  and
      King, Irwin",
    editor = "Zong, Chengqing  and
      Xia, Fei  and
      Li, Wenjie  and
      Navigli, Roberto",
    booktitle = "Proceedings of the 59th Annual Meeting of the Association for Computational Linguistics and the 11th International Joint Conference on Natural Language Processing (Volume 1: Long Papers)",
    month = aug,
    year = "2021",
    address = "Online",
    publisher = "Association for Computational Linguistics",
    url = "https://aclanthology.org/2021.acl-long.34/",
    doi = "10.18653/v1/2021.acl-long.34",
    pages = "404--414"}

@inproceedings{reimers-2019-sentence-bert,
    title = "Sentence-BERT: Sentence Embeddings using Siamese BERT-Networks",
    author = "Reimers, Nils and Gurevych, Iryna",
    booktitle = "Proceedings of the 2019 Conference on Empirical Methods in Natural Language Processing",
    month = "11",
    year = "2019",
    publisher = "Association for Computational Linguistics",
    url = "http://arxiv.org/abs/1908.10084",
}

@article{hunter2016communications,
  title={The communications gap between scientists and public: More scientists and their institutions feel a need to communicate the results and nature of research with the public},
  author={Hunter, Philip},
  journal={The EMBO Reports},
  volume={17},
  number={11},
  pages={1513--1515},
  year={2016},
  publisher={Springer}
}

@article{padt2023teaching,
  title={Teaching Public Engagement in Research Using the Engagement Tool},
  author={Padt, Frans JG and Bose, Mallika and Luloff, AE},
  journal={Journal of Planning Education and Research},
  volume={43},
  number={4},
  pages={766--772},
  year={2023},
  publisher={SAGE Publications Sage CA: Los Angeles, CA}
}

@article{kosmyna2025your,
  title={Your Brain on ChatGPT: Accumulation of Cognitive Debt when Using an AI Assistant for Essay Writing Task},
  author={Kosmyna, Nataliya and Hauptmann, Eugene and Yuan, Ye Tong and Situ, Jessica and Liao, Xian-Hao and Beresnitzky, Ashly Vivian and Braunstein, Iris and Maes, Pattie},
  journal={arXiv e-prints},
  pages={arXiv--2506},
  year={2025}
}

@inproceedings{kwon2023efficient,
  title={Efficient Memory Management for Large Language Model Serving with PagedAttention},
  author={Woosuk Kwon and Zhuohan Li and Siyuan Zhuang and Ying Sheng and Lianmin Zheng and Cody Hao Yu and Joseph E. Gonzalez and Hao Zhang and Ion Stoica},
  booktitle={Proceedings of the ACM SIGOPS 29th Symposium on Operating Systems Principles},
  year={2023}
}

@inproceedings{fick2025talk,
  title={How to talk about science in ways that are comprehensible and interesting? Evaluation of an evidence-based science communication training program for graduate students},
  author={Fick, Julian and Hendriks, Friederike and Thies, Barbara},
  booktitle={Frontiers in Education},
  volume={10},
  pages={1558203},
  year={2025},
  organization={Frontiers}
}

@article{swords2023science,
  title={Science communication training imparts confidence and influences public engagement activity},
  author={Swords, Christina M and Porter, Jerlym S and Hawkins, Amy J and Li, Edwin and Rowland-Goldsmith, Melissa and Koci, Matthew D and Tansey, John T and Woitowich, Nicole C},
  journal={Journal of microbiology \& biology education},
  volume={24},
  number={2},
  pages={e00037--23},
  year={2023},
  publisher={American Society for Microbiology 1752 N St., NW, Washington, DC}
}

@article{reincke2024identifying,
  title={Identifying focus areas for science communication training in the context of undergraduate science education},
  author={Reincke, Cathelijne M and Pieterman-Bos, Annelies and van Mil, Marc HW},
  journal={International Journal of Science Education, Part B},
  volume={14},
  number={4},
  pages={450--464},
  year={2024},
  publisher={Taylor \& Francis}
}

@inproceedings{cachola2020tldr,
  title={TLDR: Extreme Summarization of Scientific Documents},
  author={Cachola, Isabel and Lo, Kyle and Cohan, Arman and Weld, Daniel S},
  booktitle={Findings of the Association for Computational Linguistics: EMNLP 2020},
  pages={4766--4777},
  year={2020}
}

@inproceedings{cardenas2023don,
  title={‘Don’t Get Too Technical with Me’: A Discourse Structure-Based Framework for Automatic Science Journalism},
  author={Cardenas, Ronald and Yao, Bingsheng and Wang, Dakuo and Hou, Yufang},
  booktitle={Proceedings of the 2023 Conference on Empirical Methods in Natural Language Processing},
  pages={1186--1202},
  year={2023}
}

@inproceedings{vanzo2025gpt,
  title={GPT-4 as a homework tutor can improve student engagement and learning outcomes},
  author={Vanzo, Alessandro and Chowdhury, Sankalan Pal and Sachan, Mrinmaya},
  booktitle={Proceedings of the 63rd Annual Meeting of the Association for Computational Linguistics (Volume 1: Long Papers)},
  pages={31119--31136},
  year={2025}
}

@article{xu2024large,
  title={Large Language Models for Education: A Survey},
  author={Xu, Hanyi and Gan, Wensheng and Qi, Zhenlian and Wu, Jiayang and Yu, Philip S},
  journal={arXiv e-prints},
  pages={arXiv--2405},
  year={2024}
}

@inproceedings{scarlatos2025training,
  title={Training llm-based tutors to improve student learning outcomes in dialogues},
  author={Scarlatos, Alexander and Liu, Naiming and Lee, Jaewook and Baraniuk, Richard and Lan, Andrew},
  booktitle={International Conference on Artificial Intelligence in Education},
  pages={251--266},
  year={2025},
  organization={Springer}
}

@incollection{clarke2014thematic,
  title={Thematic analysis},
  author={Clarke, Victoria and Braun, Virginia},
  booktitle={Encyclopedia of critical psychology},
  pages={1947--1952},
  year={2014},
  publisher={Springer}
}

@article{asadi2020efficacy,
  title={Efficacy of masks and face coverings in controlling outward aerosol particle emission from expiratory activities},
  author={Asadi, Sima and Cappa, Christopher D and Barreda, Santiago and Wexler, Anthony S and Bouvier, Nicole M and Ristenpart, William D},
  journal={Scientific reports},
  volume={10},
  number={1},
  pages={15665},
  year={2020},
  publisher={Nature Publishing Group UK London}
}

@techreport{kincaid1975derivation,
  title={Derivation of new readability formulas (automated readability index, fog count and flesch reading ease formula) for navy enlisted personnel},
  author={Kincaid, J Peter and Fishburne Jr, Robert P and Rogers, Richard L and Chissom, Brad S},
  year={1975}
}

@inproceedings{zha2023alignscore,
  title={AlignScore: Evaluating Factual Consistency with A Unified Alignment Function},
  author={Zha, Yuheng and Yang, Yichi and Li, Ruichen and Hu, Zhiting},
  booktitle={Proceedings of the 61st Annual Meeting of the Association for Computational Linguistics (Volume 1: Long Papers)},
  pages={11328--11348},
  year={2023}
}

@inproceedings{lin2004rouge,
  title={Rouge: A package for automatic evaluation of summaries},
  author={Lin, Chin-Yew},
  booktitle={Text summarization branches out},
  pages={74--81},
  year={2004}
}
\bibliographystyle{acl_natbib}

\appendix

\section{Training Framework} \label{app:training-framework}

\paragraph{SFT Training}
We train both models, \texttt{Qwen-2.5-7b} and \texttt{Llama-3-8b}, for three epochs using the LoRA adapter with a rank of 16, an alpha value of 32, and a dropout of 0.1. The training was with a learning rate of 5e-5 and a batch size of 3. We used one NVIDIA A100-SXM4-40GB GPU, and the training took around 1 day and 23 hours for Qwen and 1 day and 4 hours for Llama.

\paragraph{DPO Training}
The DPO training of both models, Llama and Qwen, was done for one epoch, using a learning rate of 1e-5, batch size of 1, and a LoRA adapter with a rank of 16, an alpha value of 32, and a dropout of 0.1. We used one NVIDIA A100-SXM4-40GB GPU, and the training took around 1 day and 15 hours for Qwen and 1 day and 26 hours for Llama.

\begin{figure}[ht]
\begin{minipage}{.47\textwidth}

\begin{lstlisting}[basicstyle=\ttfamily\tiny, frame=single, breaklines=false, caption={Conversation Synthesis with DeepSeek}, label={lst:conv_synth_prompt},xleftmargin=0pt, xrightmargin=0pt]
Please extract the main questions answered in the 
[JOURNALISTIC-REPORT]. 
Then, simulate a conversation between a journalist and a 
researcher, where the journalist asks the extracted 
questions. For every researcher's answer, the journalist 
must ask a follow-up question to clarify unclear aspects
[SCIENTIFIC-PAPER]: <paper>
[JOURNALISTIC-REPORT]: <press release>
\end{lstlisting}

\begin{lstlisting}[basicstyle=\ttfamily\tiny, frame=single, breaklines=false, caption={Simple Journalist Role}, label={lst:journalist_role_simple},xleftmargin=0pt, xrightmargin=0pt]

You are a helpful and knowledgeable journalist asking 
questions about a scientific paper.

\end{lstlisting}

\begin{lstlisting}[basicstyle=\ttfamily\tiny, frame=single, breaklines=false, caption={Detailed Journalist Role}, label={lst:journalist_role_advanced},xleftmargin=0pt, xrightmargin=0pt]

You are a helpful and knowledgeable journalist asking 
questions about a scientific paper.
    1. Your questions encourage the researcher to place 
    their paper in a proper societal and scientific
    context to the greatest possible degree.
    
    2. Your questions focus on topics in the paper that are
    novel and have unexpected results.
    
    3. Your questions follow up on the researcher's answers,
    trying to clarify unexplained technical terms in
    everyday language.
    
Ask a single new question or a follow-up question on the
conversation. Be concise with your response. 
            
\end{lstlisting}
\end{minipage}
\label{fig:prompts}
\end{figure}
\begin{figure*}[ht]
\begin{minipage}{.99\textwidth}

\begin{lstlisting}[basicstyle=\ttfamily\small, frame=single, breaklines=false, caption={Evaluting the Researcher's response}, label={lst:societal_impact_eval_prompt},xleftmargin=0pt, xrightmargin=0pt]

Please evaluate the following text from a researcher. Identify the following:
 1. If the answer is vague and omitting important details.
 2. List of concepts that are highly technical that only an expert in the filed
 would understand. The list can be empty if the answer is not complex.

[TEXT]: {last_answer}
\end{lstlisting}

\begin{lstlisting}[basicstyle=\ttfamily\small, frame=single, breaklines=false, caption={Generating a Clarification Follow-up Question}, label={lst:follow-up-vague-answer},xleftmargin=0pt, xrightmargin=0pt]

You are a smart research journalist. Based on the paper and conversation, the last 
answer from the researcher was evaluated as follows:
  - The answer was vague.
Your task is to ask a single, concise follow-up question to address these points, 
seeking clarification and more depth. Do not be conversational, just output 
the question.

[PAPER]: {paper}
[CONVERSATION HISTORY]:{conversation}
\end{lstlisting}

\begin{lstlisting}[basicstyle=\ttfamily\small, frame=single, breaklines=false, caption={Generating a Follow-up Question on Complex Concepts}, label={lst:follow-up-complex-concepts},xleftmargin=0pt, xrightmargin=0pt]

You are a smart research journalist. Based on the paper and conversation, the last 
answer from the researcher was evaluated as follows:
  - The answer contained these complex aspects that need clarification: 
  {complex-concepts}
Your task is to ask a single, concise follow-up question to address these points,
seeking clarification and more depth. Do not be conversational, just output 
the question.

[PAPER]: {paper}
[CONVERSATION HISTORY]:{conversation}
\end{lstlisting}

\begin{lstlisting}[basicstyle=\ttfamily\small, frame=single, breaklines=false, caption={Generating a question on the Societal Impact}, label={lst:societal_impact_eval_prompt},xleftmargin=0pt, xrightmargin=0pt]
You are a smart research journalist. The last answer was clear. Now, ask a question
about the societal impact of the research paper. Do not be conversational, 
just output the question.

[PAPER]: {paper}
[CONVERSATION HISTORY]:{conversation}
\end{lstlisting}

\begin{lstlisting}[basicstyle=\ttfamily\small, frame=single, breaklines=false, caption={Generating General Question}, label={lst:societal_impact_eval_prompt},xleftmargin=0pt, xrightmargin=0pt]

You are a research journalist. Based on the paper and conversation, ask a new, 
generic question about the research. The question should not be a direct follow-up 
to the last answer. Do not be conversational, just output the question.

[PAPER]: {paper}
[CONVERSATION HISTORY]:{conversation}
\end{lstlisting}

\end{minipage}
\caption{Prompts used to generate the preference data to train our LLM Journalist models.}
\label{fig:dpo-generation-prompts}
\end{figure*}
\begin{figure}[ht]
\begin{minipage}{.45\textwidth}

\begin{lstlisting}[basicstyle=\ttfamily\tiny, frame=single, breaklines=false, caption={Extracting Scoietal Impact Questions}, label={lst:societal_impact_ques_extraction},xleftmargin=0pt, xrightmargin=0pt]

Your job is to evaluate the quality of a conversation 
between a Journalist and a Researcher by identifying
the questions that discuss the societal impact of 
the research.

Societal impact refers to the changes that occur in
people, communities, and/or environments outside of
academia as a result of the research process or
findings. These impacts can include conceptual 
changes, such as new perspectives on 
socio-ecological challenges, as well as instrumental 
and capacity-building changes that lead to 
longer-term social and environmental impacts over time. 
Stakeholder engagement plays a significant role in 
driving these societal impacts.

An example of these questions would be asking about:
    - how the research impacts society
    - real-world application benefits from this research
    - how the work improves human wellbeing
    - the negative impact of this research or any risk
    it carries
    - ethical concerns emerging from the research
    
Return the questions that discuss the societal impact in 
the following output format:
    {"high_quality_questions": []}
            
\end{lstlisting}

\begin{lstlisting}[basicstyle=\ttfamily\tiny, frame=single, breaklines=false, caption={Extracting Scientific Context Questions}, label={lst:scientific_context_ques_extraction},xleftmargin=0pt, xrightmargin=0pt]

Your job is to evaluate the quality of a conversation
between a Journalist and a Researcher, by identifying
the questions that discuss the scientific context of
the research

Scientific context puts the scientific paper in a proper 
context with respect to any other research on the same
topic, highlighting the novelty.

An example of these questions would be asking about:
    - related research on the same topic
    - how different or novel this research is in 
    comparison to previous work
    - How does this research help other scientific 
    research progress on this topic


Return the questions that discuss the scientific
context of the research in the following output 
format:
    {"high_quality_questions": []}
            
\end{lstlisting}

\begin{lstlisting}[basicstyle=\ttfamily\tiny, frame=single, breaklines=false, caption={Extracting Accessibility Questions}, label={lst:clarity_ques_extraction},xleftmargin=0pt, xrightmargin=0pt]

Your job is to evaluate the quality of a conversation 
between a Journalist and a Researcher, by identifying
the questions that ask the researcher to clarify 
complicated concepts

An example of these questions would be asking:
    - to clarify complex technical concepts that are left
    unexplained
    - to provide examples, analogies, or use descriptive
    language to understand the work
    - to provide background information that makes it 
    easy to understand the work

        

Return the questions that ask for clarification in the
following output format:
    {"high_quality_questions": []}
            
\end{lstlisting}

\end{minipage}
\caption{Prompts used to extract questions covering the different quality aspects of conversations}
\label{fig:eval-quality-prompts}
\end{figure}

\begin{figure}[ht]
\begin{minipage}{.45\textwidth}

\begin{lstlisting}[basicstyle=\ttfamily\tiny, frame=single, breaklines=false, caption={Societal Impact Evaluation}, label={lst:societal_impact_eval_prompt},xleftmargin=0pt, xrightmargin=0pt]

Your job is to evaluate the quality of a press release. 
The goal of the press release is to communicate the 
science of the paper to the public.

Societal impact refers to the changes that occur in 
people, communities, and/or environments outside of 
academia as a result of the research process or findings.
These impacts can include conceptual changes, such as 
new perspectives on socio-ecological challenges, 
as well as instrumental and capacity-building changes
that lead to longer-term social and environmental impacts
over time. Stakeholder engagement plays a significant 
role in driving these societal impacts.

Evaluate how good the press release is in placing the
paper in its proper societal context on a scale 
from 1 to 3:

    Score 1: The press release doesn't mention how the 
    research in the paper impacts society
    Score 2: The press release discusses the research 
    paper's impact on society in a very general way
    Score 3: The press release gives a very detailed 
    account of the research paper's impact on society,
    providing  examples and discussing both positive 
    and negative aspects

When evaluating a press release under this aspect, the 
following should be considered:
    - Does the press release mention how the research
    in the paper impacts society?
    - Is the discussion of the societal impact brief or 
    extensive?
    - Does the press release cover the social impact 
    of the paper in a general way, or does it mention 
    things in detail?
    - Does the press release cover only the positive 
    aspects of the paper, or does it also mention if
    the research has a negative impact?

    
First, give reasons for your score and then output the 
score in the following output format:
{"reasons": "explain your rating",  "score": "<json integer>"}
            
\end{lstlisting}

\begin{lstlisting}[basicstyle=\ttfamily\tiny, frame=single, breaklines=false, caption={Scientific Context Evaluation}, label={lst:scientific_context_eval_prompt},xleftmargin=0pt, xrightmargin=0pt]


Your job is to evaluate the quality of a press release. 
The goal of the press release is to communicate the
science of the paper to the public.

Scientific context puts the scientific paper in proper
context with respect to any other research on the same
topic, highlighting the novelty.

When evaluating a press release under this aspect, 
the following should be considered:
    - Does the press release mention related research on
    the same topic?
    - Does the press release mention the related research
    shortly or in detail?
    - Does the press release highlight how different or 
    novel this research is in comparison to previous 
    work on the topic?
    - Does the press release mention how this work helps
    other scientific research progress on this topic?


Evaluate how good the press release is in placing
the paper in its proper scientific context on a 
scale from 1 to 5:

    Score 1: The press release doesn't mention how 
    relevant the paper is to other research on 
    the topic
    Score 2: The press release mentions how relevant
    the paper is to other research on the topic in 
    a very general way
    Score 3: The press release gives a very detailed
    account of how the paper is grounded in other 
    research on the topic, highlighting the innovation
    of the paper            
First, give reasons for your score and then output the 
scorein the following output format:
    {"reasons": "explain your rating",  "score": "<json integer>"}
    
            
\end{lstlisting}

\end{minipage}
\label{fig:eval-quality-prompts-for-pr}
\end{figure}

\begin{figure}[ht]
\begin{minipage}{.45\textwidth}

\begin{lstlisting}[basicstyle=\ttfamily\tiny, frame=single, breaklines=false, caption={Language Accessibility Evaluation}, label={lst:scientific_context_eval_prompt},xleftmargin=0pt, xrightmargin=0pt]

Your job is to evaluate the quality of a press release. 
The goal of the press release is to communicate the 
science of the paper to the public.

When evaluating a press release under this aspect, the
following should be considered:

    - Does the press release contain complex 
    technical concepts that are left unexplained?
    - Does the press release provide explanations
    to clarify complex aspects of the research?
    - Does the press release clarify the research
    via examples, descriptive language, or analogies
    that help understand the work?
    - Does the press release provide background
    information that makes it easy to understand
    the paper?


Evaluate the clarity and accessibility of the press 
release language to the public on a scale 
from 1 to 5:

    Score 1: The language is very scientific and
    inaccessible to the public
    Score 2: Many aspects mentioned are very technical
    and left unexplained
    Score 3: Some aspects mentioned are technical
    and left unexplained
    Score 4: A few aspects mentioned are technical
    and left unexplained
    Score 5: The language is understandable by the
    public, and all technical aspects are clarified

    
First, give reasons for your score and then output the
score in the following output format:
    {"reasons": "explain your rating",  "score": "<json integer>"}
            
\end{lstlisting}

\end{minipage}
\label{fig:acces-eval-prompts-of-pr}
\end{figure}

\begin{figure*}
    \centering
    \includegraphics[]{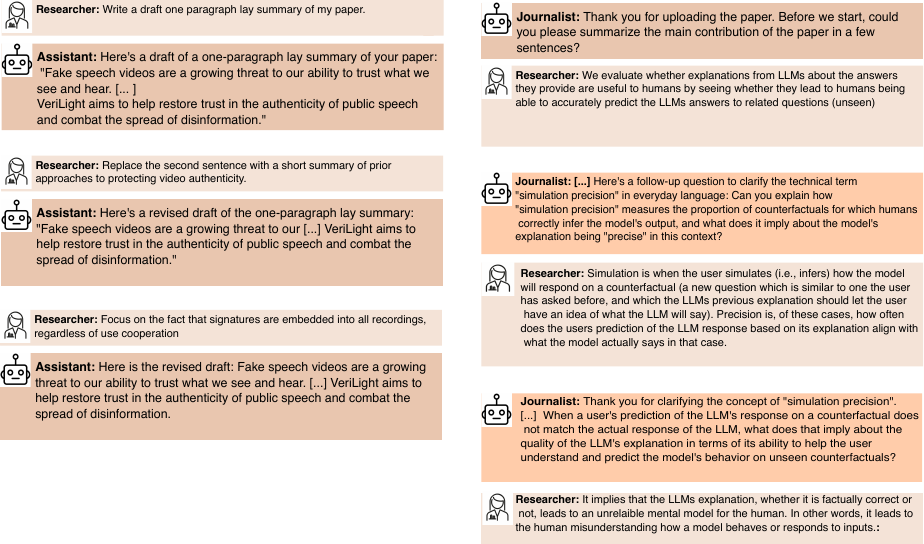}
    \caption{Example two conversations between the participants and the general-purpose LLM (left) and the LLM journalist (right).}
    \label{fig:example_conv_2}
\end{figure*}

\section{User Study}
\label{sec:user_study_questions}

\subsection{Recruitment Process}
For recruitment, we first sent out an email with a recruitment flyer on various mailing lists, asking PhD students who are interested to apply through a link to a Google Form. Next, we reached out to subjects who registered via email with more details about the user study, a link to a consent form, and a link to book an appointment slot. Lastly, during the user study, the researcher provides an introduction to the study's goal, explains the study protocol, and addresses any questions that may arise. The information sheet and Google Forms are provided in the resources of the software folder.

The user study was approved by the university's Ethics Review Board.

\subsection{Survey Questions}
The following is a list of questions the users have to answer after their interaction with each of the systems:
\begin{itemize}
    \item How mentally demanding was the task of writing the lay summary?
    \item How rushed did you feel working on the lay summary?	
    \item How successful do you think you were in writing the lay summary?
    \item How much did you experience each of the following feelings? [insecure]	
    \item How much did you experience each of the following feelings? [irritated]	
    \item How much did you experience each of the following feelings? [stressed]
    \item How much did you experience each of the following feelings? [annoyed]	
    \item Using the system helped me remember the most relevant details of the paper to highlight.	
    \item Using the system helped me use language/words for an everyday audience.	
    \item Using the system helped me frame the paper with respect to related work	
    \item Using the system helped me motivate the socieal impact of this paper.	
    \item The follow-up questions, if any, provided by the system were helpful	
    \item Please provide a brief summary describing your experience completing the lay summary for the paper.

\end{itemize}

The following is the set of questions the participants have to answer, comparing the two system interactions:
\begin{itemize}
    \item Which system do you prefer to use in the future to learn how to communicate your research to the public on your own?	
    \item My interaction with System Mango was more helpful for recalling details about the paper while writing the lay summary
    \item Which system was easier to communicate with?	
    \item Which lay summary were you more satisfied with?	
    \item How much the difference in the papers affected the output press release?
    \item Please provide a brief comparison between the two systems you interacted with
\end{itemize}

\section{Evaluation of User Study}
\begin{figure}
    \centering
    \includegraphics[width=\linewidth]{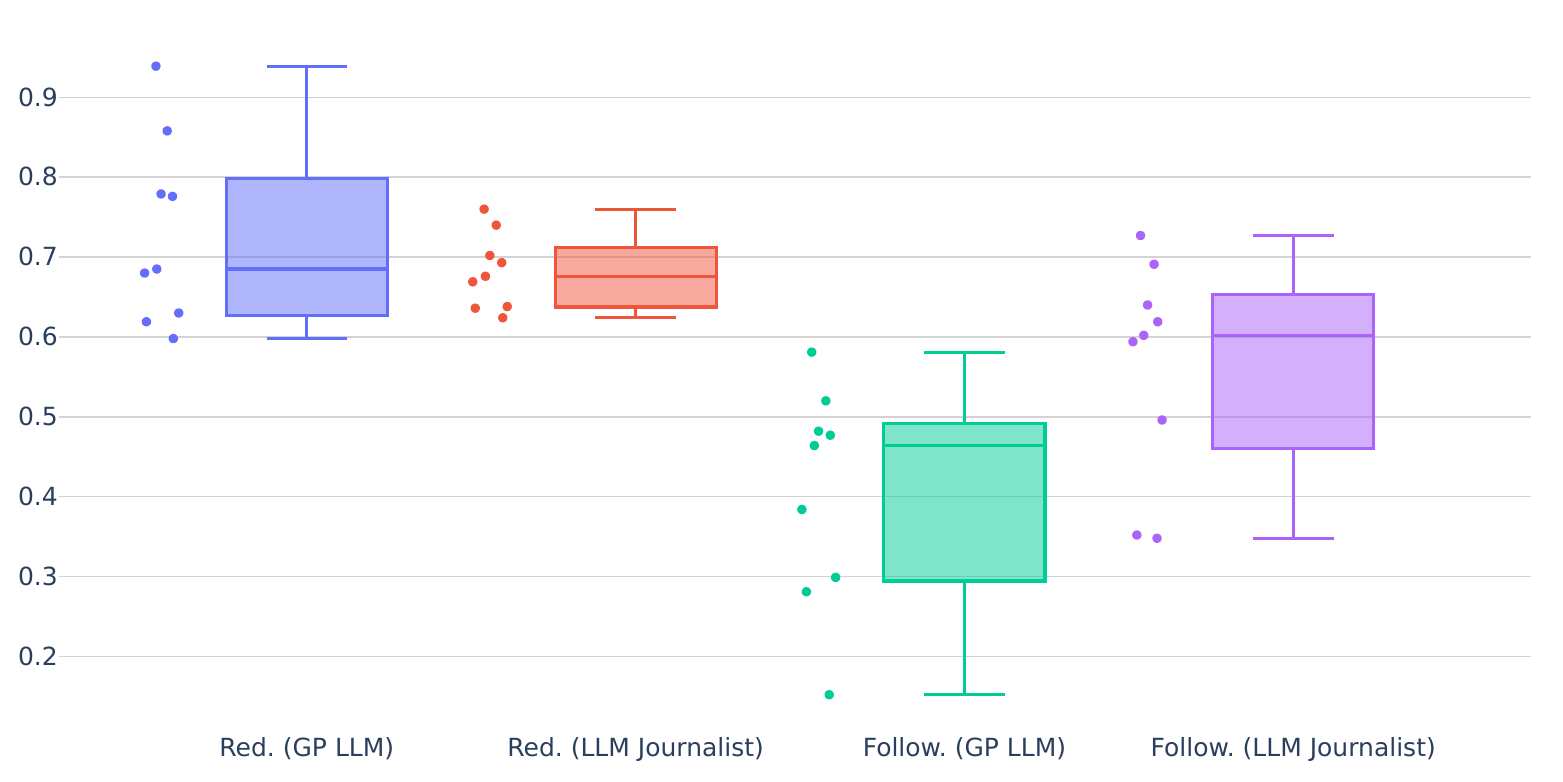}
    \caption{Automatic evaluation metric results for human study over two investigated models, general-purpose LLM (GP LLM) and the proposed LLM journalist. ``Red." denotes redundancy, and ``Follow." denotes the follow-up score.}
    \label{fig:manual_autoeval}
\end{figure}

\begin{figure}
    \centering
    \includegraphics[width=\linewidth]{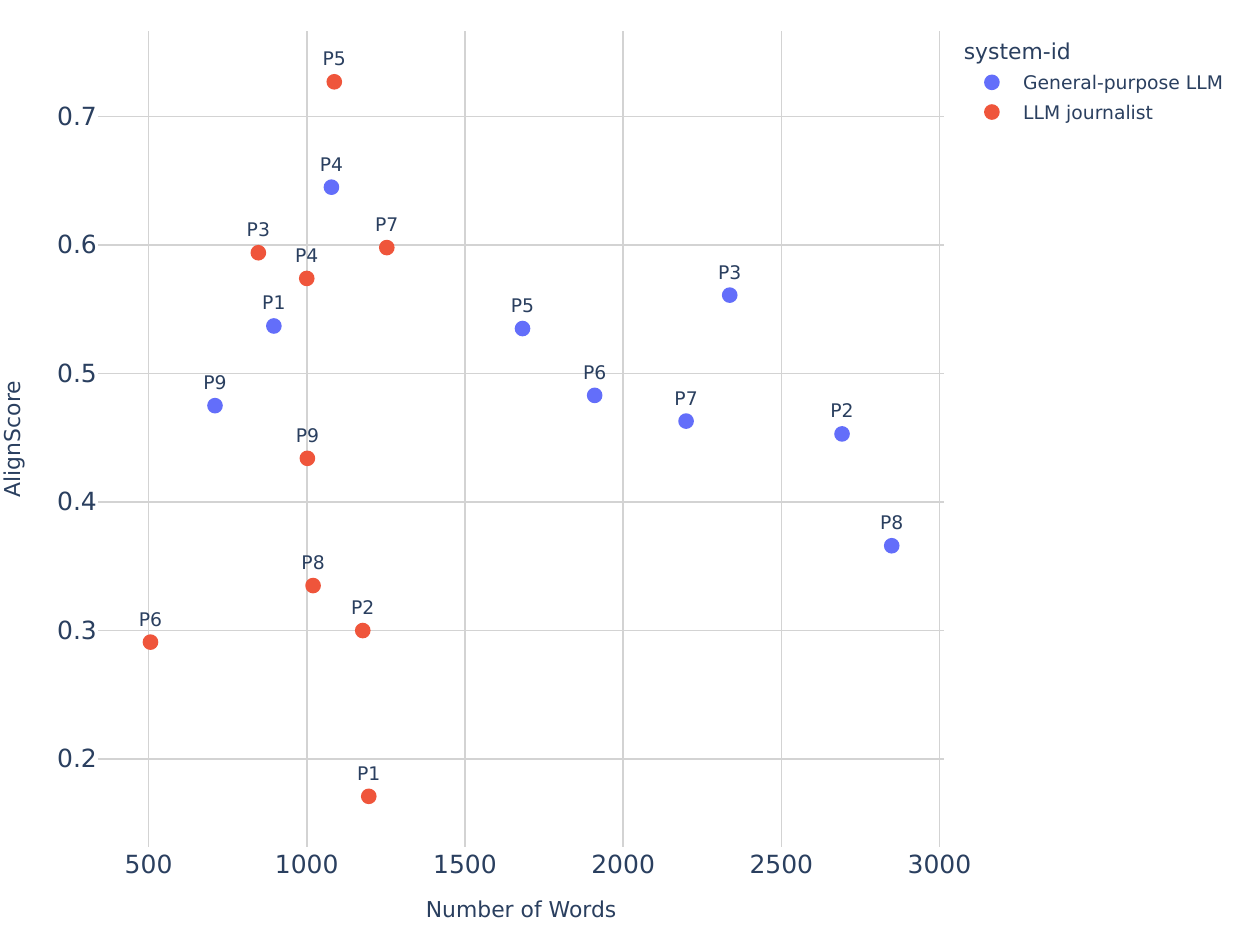}
    \caption{Correlation between AlignScore \cite{zha2023alignscore} of summary against interaction (y-axis) and the number of words in interaction (x-axis) for both evaluation settings in the human  study. }
    \label{fig:manual_numwords}
\end{figure}

Figure \ref{fig:manual_autoeval} shows the automatic evaluation results of the users' conversations with the two systems. Figure \ref{fig:manual_numwords} details the alignment score with respect to the number of words in the conversation.

\section{Qualitative Codebooks from User Study} \label{codebook}

\paragraph{Analysis of the user interactions with the assistants}
\label{sec:analysis_of_user_interactions}
To understand the patterns in user interaction with the two systems, we conduct an inductive thematic analysis of each user-issued message for both systems (LLM journalist and baseline). Across systems, there are 3 broad categories of prompt types: (1) \textit{directive}, when the user issues an instruction or action for the system to perform, (2) \textit{answering a question}, when the user responds to a question that the system posits, and (3) \textit{asking question} when the user seeks to gain information from the system (See Appendix \ref{codebook} Table \ref{tab:user-interaction-codes} for detailed code book). For the Baseline system, we found that most interactions had a majority--at least 50\% of user-issued messages--were directive prompts (6 of 9 interactions). Even when the interaction with the system was framed as a learning opportunity and that they would need to write the lay summary without assistance, users often prompted the general-purpose system with messages like ``Write a draft, one paragraph lay summary of my paper'' (P1) or ``Help me write a short summary of this paper. Go into detail on the technical parts...'' (P4). Directive messages offload the task of translating and summarizing one's research to the system, rather than doing the work themselves. 

On the contrary, for the LLM journalist system, all participants (9 of 9) had the majority of the interactions as answering questions posed by the system. This places the participant in the position to practice translating and summarizing their research. While the LLM journalist was designed to ask questions, 4 of 9 participants also issued directives like ``This is getting a bit too technical for a lay summary, can you ask me a different question?'' or questions like ``What is some advice on communicating the experiments and findings about quality filters, since these are likely not common knowledge?'' to the system demonstrating the adaptability and flexibility of the system.

\begin{table*}[t]
\centering
\caption{Affordances of LLM Journalist System}
\label{tab:affordances-llm-journalist}
\renewcommand{\arraystretch}{1.2}
\begin{tabularx}{\textwidth}{@{} p{3.2cm} X X @{}}
\toprule
\textbf{Code} & \textbf{Definition} & \textbf{Example Participant Quote} \\
\midrule

\textbf{Reducing Prompting Burden} &
The LLM system helps reduces the cognitive effort required to initiate, plan, and manage prompts during interaction, by eliminating the need to generate questions or restate constraints/task. &
``Using the [LLM journalist] system was less cognitively tiring (it prompted me with questions so I did not have to think about how to prompt the system)'' (P8)
\\
\midrule

\textbf{Scaffolding the Lay Audience} &
The degree to which the system actively or implicitly helps users adopt the perspective of a target audience (e.g., lay readers) while composing or revising content. &
``Using the [LLM journalist] system also puts you in a mindset who you are writing for.  While the [baseline] system made it harder to get in the frame of mind who you are writing for'' (P4)\\
\midrule

\textbf{Encouraging Active Engagement} &
The extent to which interaction with the system encourages deeper reasoning, reflection, and internalization of ideas, rather than passive acceptance of generated text. &
``The [LLM journalist] system was making me think deeper about the points I was making. The [baseline] system was more passive, it wasn't really making me think too much.'' (P6)
\\
\bottomrule
\end{tabularx}
\end{table*}

\begin{table*}[t]
\centering
\caption{Challenges of LLM journalist system}
\label{tab:challenges-llm-journalist}
\renewcommand{\arraystretch}{1.2}
\begin{tabularx}{\textwidth}{@{} p{3.6cm} X X @{}}
\toprule
\textbf{Code} & \textbf{Definition} & \textbf{Example Participant Quote} \\
\midrule

\textbf{Misaligned Questions} &
The LLM system asks questions that do not match what the user considers the most relevant or appropriate focus for a lay summary. &
``The [LLM journalist] system started off by focusing on the wrong aspect of the paper—maybe surprising given that the abstract highlights the societal and scientific relevance of the paper.'' (P2)
\\
\midrule

\textbf{Narrow Coverage} &
The LLM system questions address only a limited subset of dimensions needed for an effective lay summary. &
``The problem was I think they only captured one or two dimensions of the lay summary.'' (P1) \\
\midrule

\textbf{Question Redundancy} &
Repetition in LLM system questions or responses that restate prior user input without adding new value. &

``Sometimes the [LLM journalist] responses felt a little redundant, i.e., after I answered it would say something like `I agree with that' and repeat or paraphrase what I just said was important before asking the next question, which was somewhat irritating.'' (P8) \\
\bottomrule
\end{tabularx}
\end{table*}

\begin{table}[t]
\centering
\caption{User Interaction Code Book}
\label{tab:user-interaction-codes}
\begin{tabular}{p{3cm} p{6cm} p{6.5cm}}
\toprule
\textbf{Code} & \textbf{Definition} & \textbf{Example} \\
\midrule

\textbf{Directive} &
The user issues an instruction that tells the system what action to take or output to produce, regardless of surface phrasing. &
``Write a draft one paragraph lay summary of my paper.'' (P2) \\

\midrule

\textbf{Answering Question} &
The user responds to a system-posed question by supplying information, judgments, or explanations. &
Assistant: ``Could you please summarize the main contribution of the paper in a few sentences?'' \newline
User: ``The main contribution of the paper is twofold\ldots'' (P4) \\

\midrule

\textbf{Asking Question} &
The user asks the system for opinions, assessments, or perspectives. &
``Do you think that the current content could be relevant for people's daily lives?'' (P1)\\

\bottomrule
\end{tabular}
\end{table}



\end{document}